\documentclass[10pt,twocolumn,letterpaper]{article}

\usepackage{cvpr}              % To produce the CAMERA-READY version
\usepackage{graphicx}
\usepackage{xcolor}
\usepackage{array}
\usepackage[accsupp]{axessibility}  % Improves PDF readability for those with disabilities.
\usepackage{amsmath}
\usepackage{graphicx}
\usepackage[dvipsnames]{xcolor}
\usepackage{tabularx}
\usepackage{comment}
\usepackage{stfloats}
\usepackage[normalem]{ulem}
\usepackage{authoraftertitle}
\definecolor{darkgreen}{RGB}{29, 148, 27}
\definecolor{green}{RGB}{29, 148, 27}
\usepackage[pagebackref,breaklinks,colorlinks,linkcolor=blue,citecolor=blue]{hyperref}
\title{TLAC: Two-stage LMM Augmented CLIP for Zero-Shot Classification}

\author{
    Ans Munir$^{1,4}$, Faisal Z. Qureshi$^{2}$, Muhammad Haris Khan$^{3}$, Mohsen Ali$^{1}$ 
    \vspace{0.5em} \\
    \scalebox{0.85}{$^{1}$Information Technology University, Lahore, Pakistan} \\ 
    \scalebox{0.85}{$^{2}$University of Ontario Institute of Technology, Oshawa, Canada} \\ 
    \scalebox{0.85}{$^{3}$Mohamed Bin Zayed University of AI, Abu Dhabi, UAE} \hspace{0.7 cm}
    \scalebox{0.85}{$^{4}$MZASS AI} \\
    {\tt\small \{msds20033, mohsen.ali\}@itu.edu.pk, faisal.qureshi@ontariotechu.ca} \\ {\tt\small muhammad.haris@mbzuai.ac.ae}
}

\begin{document}
\maketitle
\begin{abstract}
Contrastive Language-Image Pretraining (CLIP) has shown impressive zero-shot performance on image classification. However, state-of-the-art methods often rely on fine-tuning techniques like prompt learning and adapter-based tuning to optimize CLIP's performance. The necessity for fine-tuning significantly limits CLIP's adaptability to novel datasets and domains. This requirement mandates substantial time and computational resources for each new dataset. To overcome this limitation, we introduce simple yet effective training-free approaches, Single-stage LMM Augmented CLIP (SLAC) and Two-stage LMM Augmented CLIP (TLAC), that leverages powerful Large Multimodal Models (LMMs), such as Gemini, for image classification.  The proposed methods leverages the capabilities of pre-trained LMMs, allowing for seamless adaptation to diverse datasets and domains without the need for additional training. Our approaches involve prompting the LMM to identify objects within an image. Subsequently, the CLIP text encoder determines the image class by identifying the dataset class with the highest semantic similarity to the LLM predicted object. Our models achieved superior accuracy on 9 of 11 base-to-novel datasets, including ImageNet, SUN397, and Caltech101, while maintaining a strictly training-free paradigm.  Our TLAC model achieved an overall accuracy of 83.44\%, surpassing the previous state-of-the-art few-shot methods by a margin of 6.75\%. Compared to other training-free approaches, our TLAC method achieved 83.6\% average accuracy across 13 datasets, a 9.7\% improvement over the previous methods. Our Code is available at \textcolor{magenta}{\url{https://github.com/ans92/TLAC}} %\textcolor{red}{\sout{Our method improves domain generalization, with a 3.6\% gain on ImageNetV2, 16.96\% on ImageNet-S, and 12.59\% on ImageNet-R, over prior few-shot methods.}}
% We evaluated our models on 11 base-to-novel datasets and they achieved superior accuracy on 9 of these, including benchmarks like ImageNet, SUN397 and Caltech101, while maintaining a strictly training-free paradigm.
\end{abstract}    

\section{Introduction}
%1. First describe the clip and its benefits
%2. Describe its disadvantages
%3. Then describe the advantages of multimodal
%4. And also describe the benefits of training free method.

Vision-Language Models (VLMs) like CLIP (Contrastive Language-Image Pretraining) \cite{clip} have demonstrated impressive performance on a variety of downstream tasks, including few-shot learning \cite{maple, yang2024mma} and zero-shot learning \cite{CuPL}. Trained on a massive dataset of 400 million image-text pairs using a contrastive learning approach, CLIP exhibits strong generalization capabilities. However, CLIP still necessitates a certain level of task-specific knowledge (such as fine-tuning on downstream tasks) to attain optimal results. Fine-tuning the entire model on limited downstream data is often impractical due to the risk of overfitting. To address this challenge, two primary techniques have emerged to adapt CLIP to downstream tasks: Prompt Engineering \cite{coop, cocoop} and Adapter-based Fine-tuning \cite{clip-adapter, chen2022adaptformer}.

CLIP rely on textual prompts to guide its image understanding. When presented with an image, CLIP aims to identify the most relevant prompt from a vast pool of possibilities. For instance, given an image of a Red Car, CLIP might match it with the prompt ``A photo of a red car.'' Previous work \cite{coop} has shown that changes in the structure of the CLIP prompt affect its accuracy. To address this challenge and to adapt CLIP to specific downstream tasks without extensive fine-tuning, prompt learning has emerged as a promising technique. Through training, the model learns to exploit the fixed elements within prompts such as ``\emph{A photo of a/an} [object],'' allowing it to effectively apply this knowledge to downstream tasks. Another approach leverages lightweight, trainable adapter networks, leaving the larger CLIP model frozen. These techniques, in conjunction with CLIP, offer strong generalization capabilities and have achieved impressive results. However, a major limitation of these methods is the need for training on each new dataset. This hinders their potential as universal models capable of handling diverse domains and classes. 

To mitigate these challenges, recent training-free approaches, such as CuPL \cite{CuPL} and MPVR \cite{MPVR}, integrate Large Language Models (LLMs) with CLIP. Specifically, these methods first employ LLMs to generate textual descriptions for each class. Then, CLIP is used to determine the class description that best aligns with the visual representation of the image. While eliminating the need for training, these methods still necessitate the generation of class descriptions tailored to each new dataset.

To address this limitation, we propose a straightforward technique that leverages the capabilities of Large Multimodal Models (LMMs) \footnote{In literature some works called them Multimodal Large Language Models (MLLMs)} such as Gemini \cite{team2023gemini} eliminating the need for explicit class description generation. Our approach is training-free, allowing us to fully harness the power of LMMs, which benefit from extensive pre-training on diverse data. A key advantage of LMMs is their universality, enabling them to handle a wide range of datasets and domains. To the best of our knowledge, this is the first work to explore LMM for image classification and combined LMM, Gemini, with VLM, CLIP. The main contribution of this work is as follows:

\begin{itemize}
    \item We propose a simple yet effective training-free approach to leverage Large Multimodal Models (LMMs) for image classification task. 
    \item Specifically, this work introduces two models, Single-stage LMM Augmented CLIP (SLAC) and Two-stage LMM Augmented CLIP (TLAC), which combine the strengths of LMM and VLM to leverage their combined capabilities.
    \item Our experiments demonstrate that our approach achieves superior accuracy on a majority of evaluated datasets, including the large-scale ImageNet, all while remaining entirely training-free and requiring no training samples.
\end{itemize}

\begin{figure}[t]
\begin{center}
   \includegraphics[width=1.0\linewidth]{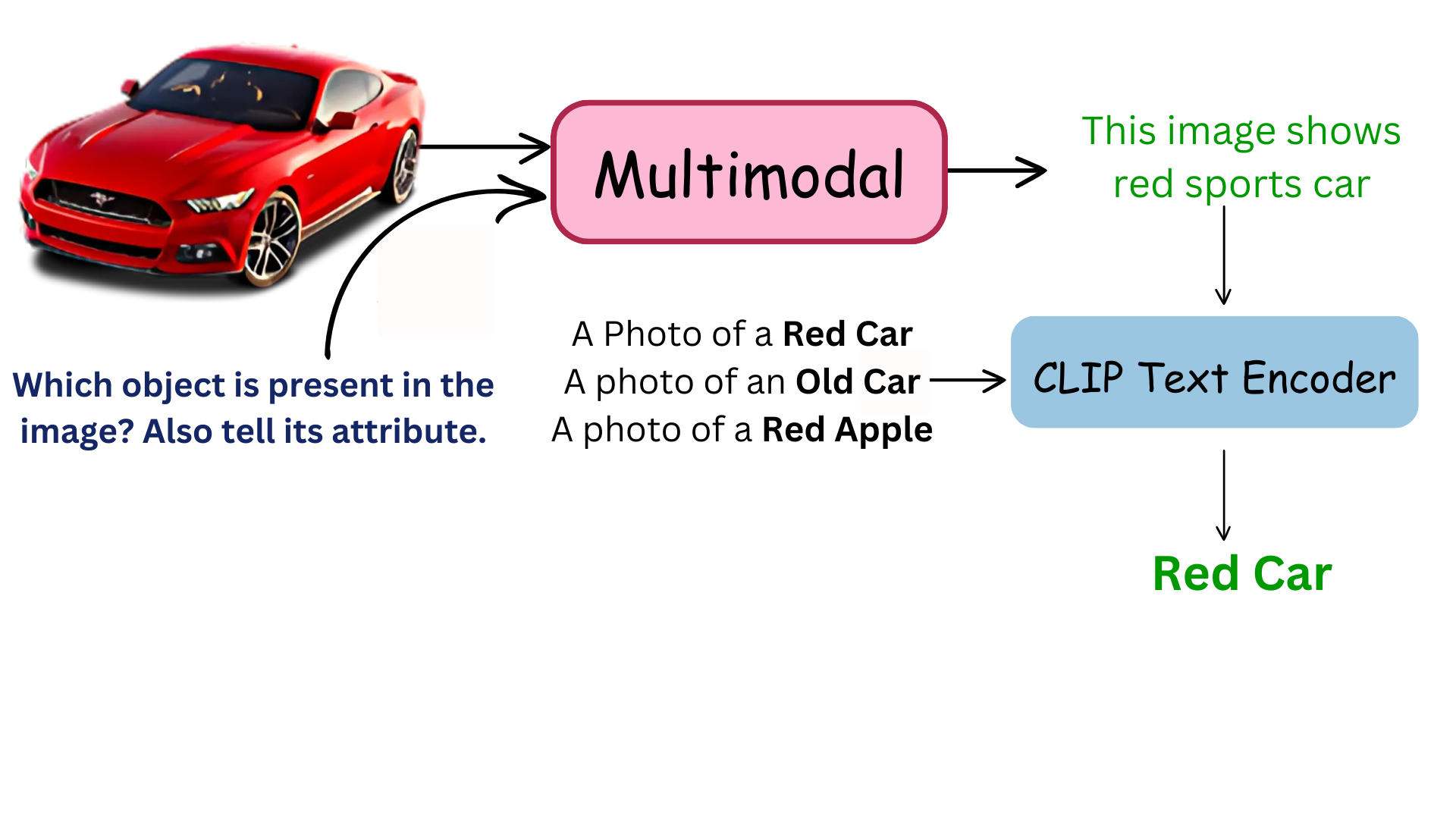}
\end{center}
   \caption{Diagram of SLAC model. Blue text is the LMM prompt. Red Car, Old Car and Red Apple are dataset class labels.}
\label{fig:llama-diagram}
\end{figure}

\begin{figure}[]
\begin{center}
   \includegraphics[width=1.0\linewidth]{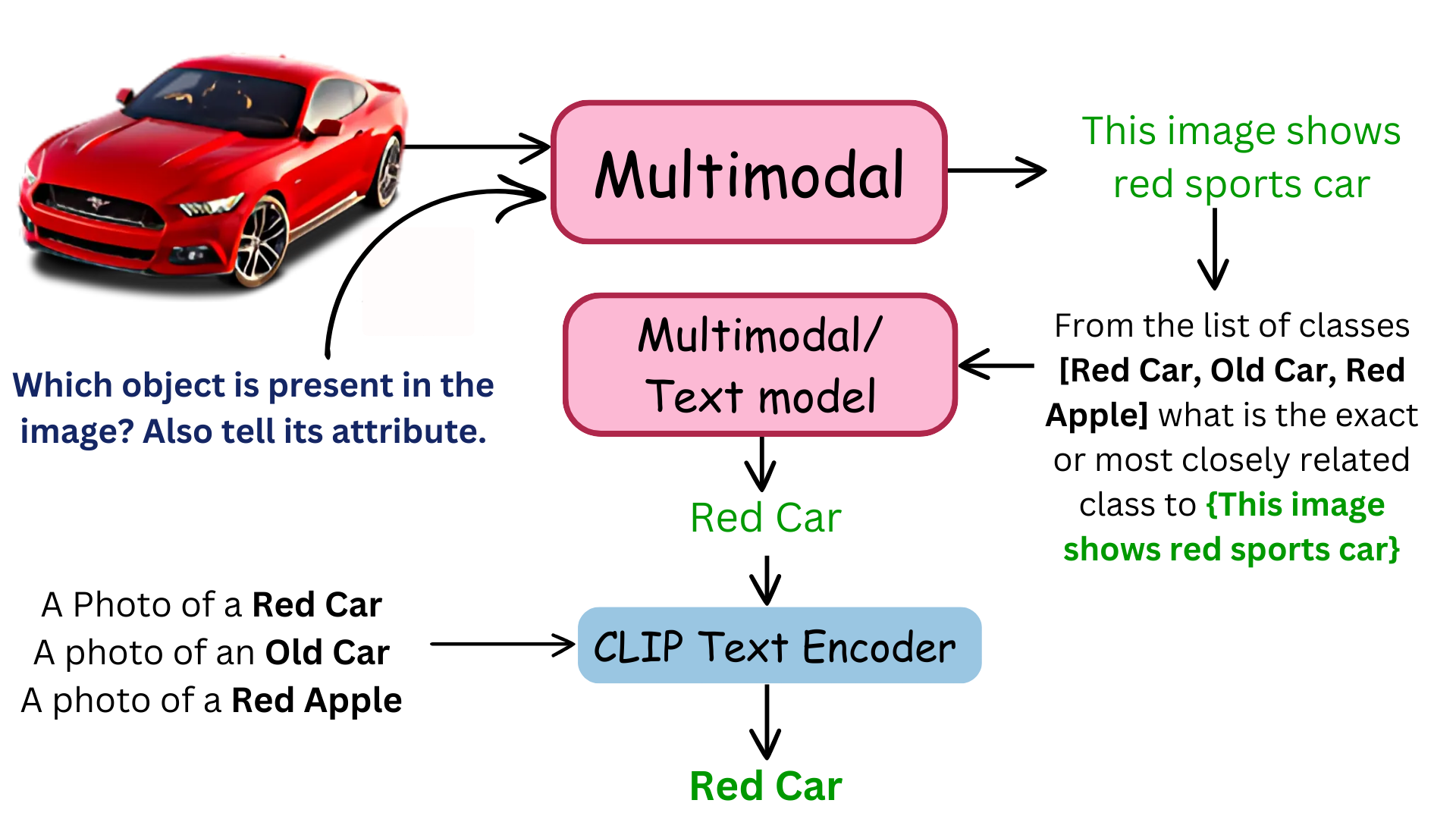}
\end{center}
   \caption{Diagram of our TLAC model. Blue text is the LMM prompt. Red Car, Old Car and Red Apple are dataset class labels.}
\label{fig:2-step-approach}
\end{figure}

\section{Method}
Unlike previous work \cite{maple, yang2024mma, coop, cocoop} that relied on fine-tuning CLIP, we propose a  training-free approach leveraging Large Multimodal Models (LMMs) for image classification. Our method offers a significant advantage: it's training-free, yet surpasses the performance of state-of-the-art fine-tuned models. Furthermore, our method requires no training samples, unlike some other training-free approaches \cite{zhang2022tip-adapter}. We introduce two distinct models: SLAC and TLAC. 

\subsection{Single-stage LMM Augmented CLIP (SLAC)}
In our SLAC model, we first provide an image and a prompt to the LMM, Gemini. For instance, the prompt might be \emph{``Which object is present in the image? Also tell its attribute.''} Gemini then generates a textual description, $z$, such as \emph{``The image shows red sports car''} as illustrated in Figure \ref{fig:llama-diagram}. Subsequently, both $z$ and all dataset class labels, $y$, are encoded into textual features using the CLIP text encoder. Finally, the similarity between $z$ and each $y$ is computed to determine the image's predicted class.

\begin{equation}
    c^* = \underset{y\in C}{\arg\max} \Bigl( g(z) . g(y) \Bigr)
\end{equation}
where $z$ is the answer predicted by Gemini LMM and $y$  represents the class from the dataset classes $C$.

\subsection{Two-stage LMM Augmented CLIP (TLAC)}
Our SLAC model yielded promising results. However, we encountered instances where the model correctly identified the class present in the image, but the predicted label did not match the ground truth. For example, in flower datasets, some images were labeled with common names, while others used scientific names. Although our SLAC model, incorporating CLIP, mitigated this issue to some extent, it still struggled with scientific names. To address this limitation, we propose a TLAC model, as illustrated in Figure \ref{fig:2-step-approach}.

TLAC model employs a two-step approach. First, the LMM, Gemini, gives the answer to our question with respect to the image, similar to the SLAC model. However, instead of predicting the object present in the image, we ask the LMM to provide the object most relevant to the dataset class, based on its initial prediction. If the initial prediction was accurate, the second step reinforces it. However, if the initial prediction was partially correct but used a different name (for example, \emph{Gaillardia} instead of \emph{Blanket Flower} as shown in Table \ref{tab:images-2step}), the second step corrects and improves the prediction, leading to higher accuracy. Figure \ref{fig:2-step-approach} provides a visual overview of this TLAC model.

%%%%%%%%%%%%%%%% Training Free Table %%%%%%%%%%%%%%%
\begin{table*}[]
    % \centering
    \begin{tabular}{|c|c|c|c|c|c|c|c|c|c|c|c|c|c|c|} \hline
  & \rotatebox{90}{ImageNet} & \rotatebox{90}{Caltech101} & \rotatebox{90}{OxfordPets} & \rotatebox{90}{StanfordCars} & \rotatebox{90}{Flowers102} & \rotatebox{90}{Food101} & \rotatebox{90}{FGVCAircraft} & \rotatebox{90}{SUN397} & \rotatebox{90}{DTD} & \rotatebox{90}{UCF101} & \rotatebox{90}{ImageNetv2} & \rotatebox{90}{ImageNet-R} & \rotatebox{90}{ImageNet-S} & \rotatebox{90}{Average} \\ \hline

  CLIP-S & 73.5 & 94.3 & 93.1 &  76.9 & 76.2 & 90.3 & 30.0 & 67.6 &  52.5 & 73.8 & 60.9 & 74.0 & 46.2 & 69.9 \\

  CLIP-DS &  75.5 & 93.7 & 93.5 & 78.1 & 79.5 & 90.9 & 31.8 & 69.0 & 54.8 & 76.2 & 61.9 & 77.7 & 48.8 & 71.6 \\

  %CuPL & 76.7 & 96.1 & 94.3 & 64.2 & 79.6 & 91.4 & 35.1 & 72.8 & 61.1 & 75.8 & 95.8 & 78.6 & - \\

  % Tip-Adapter &  62.01 & 90.18 & 88.14 & 66.77 & 89.89 & 77.83 & 29.76 & 66.85 & 60.93 & 70.58 & - & - & - \\
  
  CuPL & \textcolor{blue}{76.7} & 93.5 & 93.8 & 77.6 & 79.7 &  \textcolor{blue}{93.4} & 36.1 & 73.3 & 61.7 & 78.4 &  63.4 & - & - & 75.2 \\ 

  D-CLIP & 75.1 & \textbf{97.0} & 93.0 &  75.1 & 79.5 & 91.1 & 31.8 &  69.6 & 56.1 & 76.2 & 62.2 & 76.5 & 48.9 & 71.7 \\

  Waffle & 75.1 & 96.2 & 93.2 & 76.5 & 78.3 & 91.5 & 32.5 &  69.4 & 55.3 & 76.0 & 62.3 & 77.0 & 49.1 & 71.7 \\

  MPVR (Mix) & 75.9 & 95.4 & 93.1 & 70.6 & \textcolor{blue}{83.8} & 91.4 & 37.6 & 72.5 & 61.6 & 75.8 & 62.2  & 78.4  & 49.7 & 72.9  \\

  MPVR (GPT) & \textbf{76.8} & 96.1 & 93.7 & 78.3 & 83.6 & 91.5 & 34.4 & 73.0 &  \textcolor{blue}{62.9} & 78.1 & 63.4 & 78.2 & 50.6 & 73.9 \\

  % SuS-X & 70.00 & 93.96 & 91.58 & 66.13 & 73.81 & 86.08 & 30.51 & 67.85 & 54.55 & 66.72 & 90.94 & 68.66 & 47.71 \\
  
\hline
  \textbf{Ours (SLAC)} & 73.8 & \textcolor{blue}{96.6} & \textcolor{blue}{96.5} & \textcolor{blue}{88.7} & 77.7 & 92.9 & \textcolor{blue}{65.6} & \textcolor{blue}{73.5} & 58.5 & \textcolor{blue}{85.2} & \textcolor{blue}{67.9} & \textcolor{blue}{89.9} & \textcolor{blue}{66.1} & \textcolor{blue}{79.4} \\ 

  \textbf{Ours (TLAC)} & 74.1 & \textbf{97.0} & \textbf{97.1} & \textbf{90.2} & \textbf{85.7} & \textbf{94.4} & \textbf{79.4} & \textbf{79.0} & \textbf{72.6} & \textbf{89.5} & \textbf{69.2} & \textbf{90.8} & \textbf{68.2} & \textbf{83.6} \\ \hline
\end{tabular}
    \caption{Table compares the results of our models with those of previous training-free methods. Results of previous state-of-the-art models have been taken from \cite{MPVR}.The best result is displayed in bold, while the second-highest result is shown in blue. Higher scores represent superior performance.}
    \label{tab:training-free-methods}
\end{table*}
%%%%%%%%%%%%%%%%%%% End of Training-free table %%%%%%%%%%%%%%

\section{Experiments}

\subsection{Generalization from Base-to-Novel classes}
Previous methods \cite{maple, yang2024mma, coop, cocoop} have been trained on base classes for 16 instances per class in a few-shot manner also known as 16-shot learning and then generalized the knowledge on novel classes. In contrast, our approach is training-free, making it applicable to new datasets and domains. Similar to prior work \cite{maple, yang2024mma, coop, cocoop}, we evaluated our approach on 11 diverse image classification datasets such as ImageNet \cite{imagenet} and Caltech101 \cite{caltech101}, 2 general object classification datasets; OxfordPets \cite{oxfordpets}, StanfordCars \cite{stanfordcars}, Flowers102 \cite{flowers102}, Food101 \cite{food101}, and FGVCAircraft \cite{fgvcaircraft}, 5 fine-grained datasets; SUN397 \cite{sun397}, a scene understanding dataset; DTD, a satellite-image recognition dataset \cite{dtd}; UCF101 \cite{ucf101}, an action classification dataset.

\subsection{Domain Generalization}
In domain generalization, models are evaluated on out-of-distribution datasets to assess their robustness across different domains. \cite{cocoop} proposed testing ImageNet-trained models on ImageNet variants such as ImageNetV2 \cite{ImageNetV2}, ImageNet-Sketch \cite{ImageNet-Sketch}, and ImageNet-R \cite{ImageNet-R}. Results \ref{tab:domain-gen} demonstrate that our approach achieves superior results on these diverse domains, even without explicit training. %Table \ref{tab:domain-generalization-images} in the supplementary material provides visual examples of these domain generalization datasets.

\subsection{Implementation Details}
For the task of image classification we employed two versions of Gemini: Gemini Flash 1.5-002 and Gemini Pro 1.5-002 as our Large Multimodal Model (LMM). In SLAC model, we mostly used Gemini Pro, while in second step of TLAC model, we used Gemini Flash. Consistent with prior work \cite{maple, yang2024mma}, we utilized the CLIP-B/16 model as a VLM text encoder.

%%%%%%%%%%%%%%%%%%%%% Few-shot Table %%%%%%%%%%%%%%%%%%%

\begin{table*}[t]

\begin{center}
    
  {\renewcommand{\arraystretch}{1.4}% for the vertical padding  
  \begin{tabular}{c|c c|c c|c c|c c|c c|c c} \toprule
    &  \multicolumn{2}{c|}{Overall Avg} &  \multicolumn{2}{c|}{ImageNet} &  \multicolumn{2}{c|}{Caltech101} & \multicolumn{2}{c|}{OxfordPets} &  \multicolumn{2}{c|}{StanfordCars} &  \multicolumn{2}{c}{Flowers102} \\
    Method & Base & Novel & Base & Novel & Base & Novel & Base & Novel & Base & Novel & Base & Novel \\ \hline

    CLIP & 69.34 & 74.22 & 72.43 & 68.14 & 96.84 & 94.00 & 91.17 & 97.26 & 63.37 & 74.89 & 72.08 & 77.80 \\

    CoOp & 82.69 & 63.22 & 76.47 & 67.88 & 98.00 & 89.81 & 93.67 & 95.29 & 78.12 & 60.40 & 97.60 & 59.67 \\

    Co-CoOp & 80.47 & 71.69 & 75.98 & 70.43 & 97.96 & 93.81 & 95.20 & 97.69 & 70.49 & 73.59 & 94.87 & 71.75 \\

    ProDA & 81.56 & 72.30 & 75.40 & 70.23 & 98.27 & 93.23 & 95.43 & 97.83 & 74.70 & 71.20 & 97.70 & 68.68 \\

    KgCoOp & 80.73 & 73.60 & 75.83 & 69.96 & 97.72 & 94.39 & 94.65 & 97.76 & 71.76 & 75.04 & 95.00 & 74.73 \\

    MaPLe & 82.28 & 75.14 & 76.66 & 70.54 & 97.74 & 94.36 & 95.43 & 97.76 & 72.94 & 74.00 & 95.92 & 72.46 \\

    LASP & 82.70 & 74.90 & 76.20 & 70.95 & 98.10 & 94.24 & 95.90 & \textcolor{blue}{97.93} & 75.17 & 71.60 & 97.00 & 74.00 \\

    % \textcolor{silver}{LASP-V} & \textcolor{silver}{83.18} & \textcolor{silver}{76.11} & \textcolor{silver}{76.25} & \textcolor{silver}{71.17} & \textcolor{silver}{98.17} & \textcolor{silver}{94.33} & \textcolor{silver}{95.73} & \textcolor{silver}{97.87} & \textcolor{silver}{75.23} & \textcolor{silver}{71.77} & \textcolor{silver}{97.17} & \textcolor{silver}{73.53} \\ 

    RPO & 81.13 & 75.00 & 76.60 & 71.57 & 97.97 & 94.37 & 94.63 & 97.50 & 73.87 & 75.53 & 94.13 & 76.67 \\

    MMA & 83.20 & 76.80 & 77.31 & 71.00 & 98.40 & 94.00 & 95.40 & \textbf{98.07} & 78.50 & 73.10 & 97.77 & 75.93 \\ \hline 

    \textbf{Ours (SLAC)} & 74.43 & \textcolor{blue}{78.69} & 79.90 & \textcolor{blue}{73.78} & 94.19 & \textcolor{blue}{96.62} & 93.62 & 96.48 & 74.79 & \textcolor{blue}{88.73} & 77.21 & \textcolor{blue}{77.73} \\

    % Ours (Gem F) &- & 76.31 &- & 74.04 & 94.54 & 95.63 & 92.13 & 96.31 & 73.49 & 90.20 & 79.96 & 76.6 \\ 
    
    \textbf{Ours (TLAC)} & 76.81 & \textbf{83.44} & 80.10 & \textbf{74.06} & 91.67 & \textbf{96.96} & 94.26 & 97.09 & 74.99 & \textbf{90.17} & 79.01 & \textbf{85.74} \\ 
     
     & & \textcolor{green}{+6.75} &  & \textcolor{green}{+2.49} &  & \textcolor{green}{+2.57} &  & \textcolor{red}{-0.98} &  & \textcolor{green}{+14.64} &  & \textcolor{green}{+9.07} \\
    \hline
    
  \end{tabular}
\par\vspace{0.2cm}\par
  \begin{tabular}{c|c c|c c|c c|c c|c c|c c}  \hline
    &  \multicolumn{2}{c|}{Food101} &  \multicolumn{2}{c|}{FGVCAircraft} &  \multicolumn{2}{c|}{SUN397} & \multicolumn{2}{c|}{DTD} &  \multicolumn{2}{c|}{EuroSAT} &  \multicolumn{2}{c}{UCF101} \\
    Method & Base & Novel & Base & Novel & Base & Novel & Base & Novel & Base & Novel & Base & Novel \\ \hline

    CLIP & 90.10 & 91.22 & 27.19 & 36.29 & 69.36 & 75.35 & 53.24 & 59.90 & 56.48 & 64.05 & 70.53 & 77.50 \\

    CoOp & 88.33 & 82.26 & 40.44 & 22.30 & 80.60 & 65.89 & 79.44 & 41.18 & 92.19 & 54.74 & 84.69 & 56.05 \\

    Co-CoOp & 90.70 & 91.29 & 33.41 & 23.71 & 79.74 & 76.86 & 77.01 & 56.00 & 87.49 & 60.04 & 82.33 & 73.45 \\

    ProDA & 90.30 & 88.57 & 36.90 & 34.13 & 78.67 & 76.93 & 80.67 & 56.48 & 83.90 & 66.00 & 85.23 & 71.97 \\

    KgCoOp & 90.50 & 91.70 & 36.21 & 33.55 & 80.29 & 76.53 & 77.55 & 54.99 & 85.64 & 64.34 & 82.89 & 76.67\\

    MaPLe & 90.71 & 92.05 & 37.44 & 35.61 & 80.82 & \textcolor{blue}{78.70} & 80.36 & 59.18 & 94.07 & 73.23 & 83.00 & 78.66 \\

    LASP &  91.20 & 91.70 & 34.53 & 30.57 & 80.70 & 78.60 & 81.40 & 58.60 & 94.60 & \textcolor{blue}{77.78} & 84.77 & 78.03 \\

    % \textcolor{silver}{LASP-V} & \textcolor{silver}{91.20} & \textcolor{silver}{91.90 } & \textcolor{silver}{38.05} & \textcolor{silver}{33.20} & \textcolor{silver}{80.70} & \textcolor{silver}{79.30} & \textcolor{silver}{81.10} & \textcolor{silver}{62.57} & \textcolor{silver}{95.00} & \textcolor{silver}{83.37} & \textcolor{silver}{85.53} & \textcolor{silver}{78.20} \\ 

    RPO & 90.33 & 90.83 & 37.33 & 34.20 & 80.60 & 77.80 & 76.70 & 62.13 & 86.63 & 68.97 & 83.67 & 75.43\\

    MMA & 90.13 & 91.30 & 40.57 & 36.33 & 82.27 & 78.57 & 83.20 & \textcolor{blue}{65.63} & 85.46 & \textbf{82.34} & 86.23 & 80.03\\ \hline 

    \textbf{Ours (SLAC)} & 92.51 & \textcolor{blue}{92.87} & 56.42 & \textcolor{blue}{65.60} & 69.89 & 73.45 & 52.08 & 58.45 & 48.78 & 56.72 & 79.37 & \textcolor{blue}{85.18} \\ % food101-pro(89.82,91.62) 

    % Ours (Gem F) & \textbf{92.51} & \textbf{92.87} & 53.42 & 65.3 & - & 74.06 & - & 56.76 & - & 47.9 & 79.01 & 83.99 \\ 

    \textbf{Ours (TLAC)} & 88.46 & \textbf{94.37} & 62.79 & \textbf{79.36} & 73.32 & \textbf{79.00} & 68.06 & \textbf{72.95} & 49.10 & 58.67 & 83.13 & \textbf{89.51} \\ 
    
    & & \textcolor{green}{+2.32} &  & \textcolor{green}{+43.03} &  & \textcolor{green}{+0.30} &  & \textcolor{green}{+7.32} &  & \textcolor{red}{-23.67}  &  & \textcolor{green}{+9.48} \\
    \hline
    
  \end{tabular}
  }
  \caption{Table shows the results of previous state-of-the-art few-shot methods on 11 base-to-novel datasets. Base accuracy is on training/seen classes while Novel accuracy is on new/unseen classes. Results of both of our models, SLAC and TLAC are also shown. Previous state-of-the-art results have been taken from \cite{yang2024mma}. Although not directly comparable to prior base accuracy due to our no-training approach, we still mention base accuracy to facilitate comparison between prior results that require training and our training-free method. Results show that our models demonstrate superior accuracy on novel classes as compared to fine-tuned models. The best result is displayed in bold, while the second-highest result is shown in blue. Higher results are better.}
  \label{tab:fewshot-dataset-results}
  \end{center}
\end{table*}

%%%%%%%%%%%%%% End of few-shot table %%%%%%%%%%%%%%%

\subsection{Main Results}

\subsection*{Training-Free Methods}
We compared the performance of our model against several training-free methods, including CLIP \cite{clip} (using both a simple prompt, CLIP-S, and dataset-specific prompts, CLIP-DS), CuPL \cite{CuPL}, D-CLIP \cite{D-Clip}, Waffle \cite{waffle}, and MPVR \cite{MPVR}. Table \ref{tab:training-free-methods} presents the results, showing that our model achieved superior accuracy on all datasets except ImageNet. Our SLAC and TLAC models achieved overall average accuracies of 79.4\% and 83.6\%, respectively, surpassing the previous highest accuracy of 73.9\% reported by MPVR \cite{MPVR} with the margin of 9.7\%.

On ImageNet, TLAC achieved 74.1\%, slightly below the state-of-the-art results of MPVR (76.8\%) and CuPL (76.7\%). However, on Caltech101, TLAC reached 97.0\%, matching the performance of D-CLIP. Notably, TLAC surpassed the previous best on OxfordPets, achieving 97.1\%, a 3.3\% improvement over CuPL's 93.8\%. Significant accuracy gains were observed on StanfordCars (11.1\%), FGVCAircrafts (41.8\%), SUN397 (5.7\%), and DTD (9.7\%). Furthermore, TLAC demonstrated superior domain generalization, achieving improvements of 5.8\% on ImageNetv2, 12.4\% on ImageNet-R, and 17.6\% on ImageNet-S.

\subsection*{Base-to-Novel Generalization}
In this experimental setup, we compared our approach with several state-of-the-art few-shot methods: CLIP \cite{clip}, CoOp \cite{coop}, Co-CoOp \cite{cocoop}, ProDA \cite{proDA}, KgCoOp \cite{KgCoOp}, MaPLe \cite{maple}, LASP \cite{lasp}, RPO \cite{RPO}, and MMA \cite{yang2024mma}. These methods typically train on base classes in a 16-shot learning manner. Table \ref{tab:fewshot-dataset-results} presents the results for base class accuracy (\textbf{Base}) and novel class accuracy (\textbf{Novel}). Our approach, which does not involve training or fine-tuning, prevents direct comparison with the \textbf{Base} results of prior methods. Nevertheless, results \ref{tab:fewshot-dataset-results} demonstrate that our model is competitive with fine-tuned models in terms of Base accuracy while achieving higher Novel accuracy. On three datasets-ImageNet, Food101, and FGVCAircraft-our model achieved higher Base accuracy than previous fine-tuned models, while remaining comparable on other datasets. This highlights the competitiveness of our model compared to other fine-tuned models, even on seen classes.

\begin{figure}[t]
\begin{center}
   \includegraphics[width=1.0\linewidth]{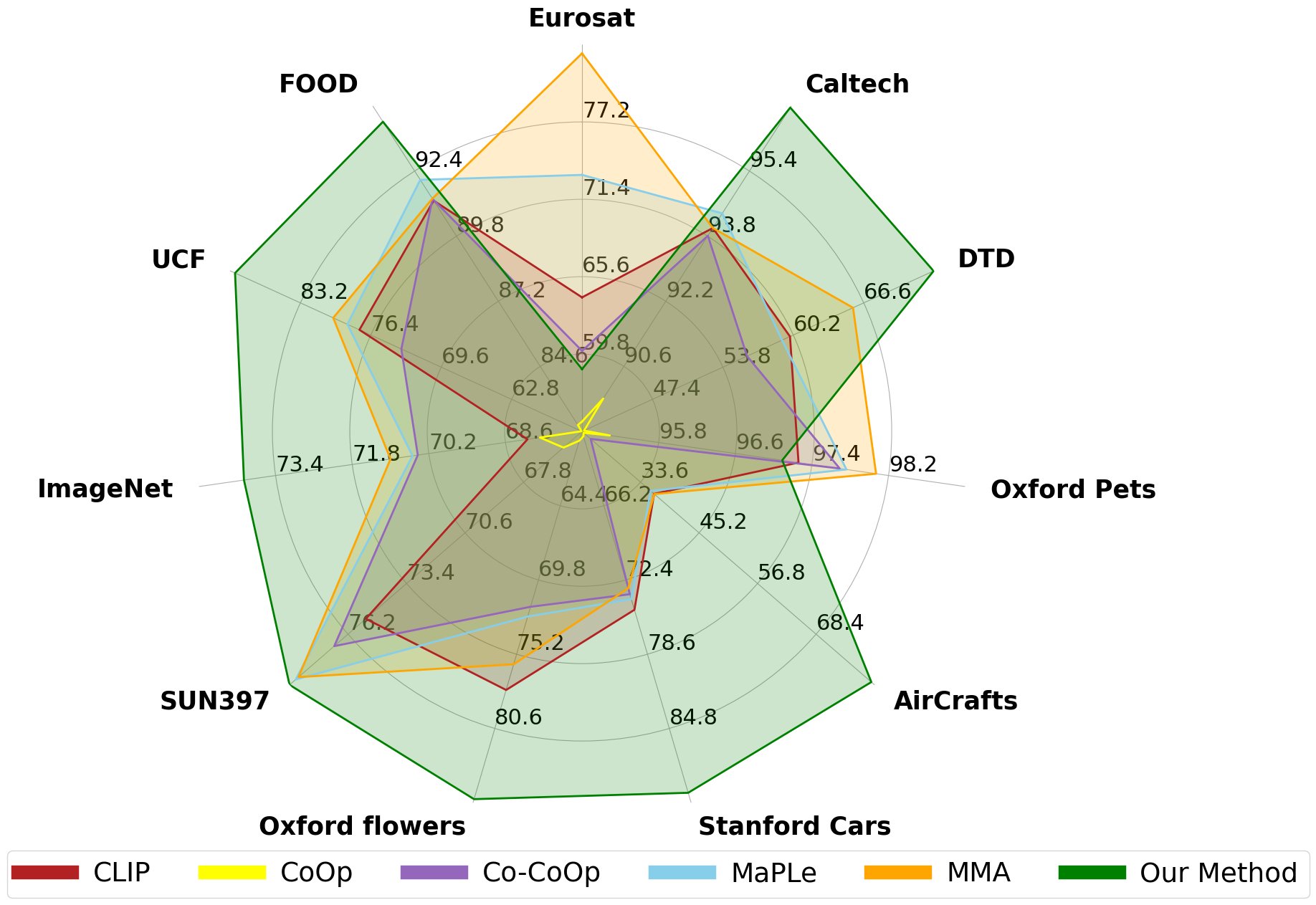}
\end{center}
   \caption{The figure illustrates the superior performance of our model compared to previous state-of-the-art few-shot methods on novel classes. Our approach has demonstrated significant performance gains across multiple challenging datasets, including Stanford Cars, Aircrafts, Oxford Flowers, ImageNet, and Caltech.}
\label{fig:spider-chart}
\end{figure}

 Our TLAC model achieved the highest overall accuracy of 83.44\%, surpassing the previous state-of-the-art, MMA \cite{yang2024mma}, by 6.75\%. On general object recognition datasets, our approach outperformed state-of-the-art methods by 2.49\% on ImageNet and 2.57\% on Caltech101 in novel accuracy. For fine-grained image recognition, we achieved significant improvements of 14.64\%, 9.07\%, 2.32\% and 43.03\% on StanfordCars, Flowers102, Food101 and FGVCAircraft, respectively. However, we underperformed on OxfordPets by 0.98\%. Additionally, we observed better performance on DTD and UCF101, with improvements of 7.32\% and 1.04\%, respectively. We also achieved better accuracy on scene understanding dataset, SUN397. On EuroSAT, our accuracy fell short of previous models. Figure \ref{fig:spider-chart} presents a comparison of the overall performance of our model with that of previous state-of-the-art models.

\begin{table}[]
    % \centering
    \begin{tabular}{c c c c}
    \toprule
        \includegraphics[width=48px, height=48px]{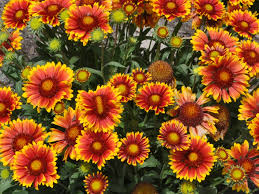} & \includegraphics[width=48px, height=48px]{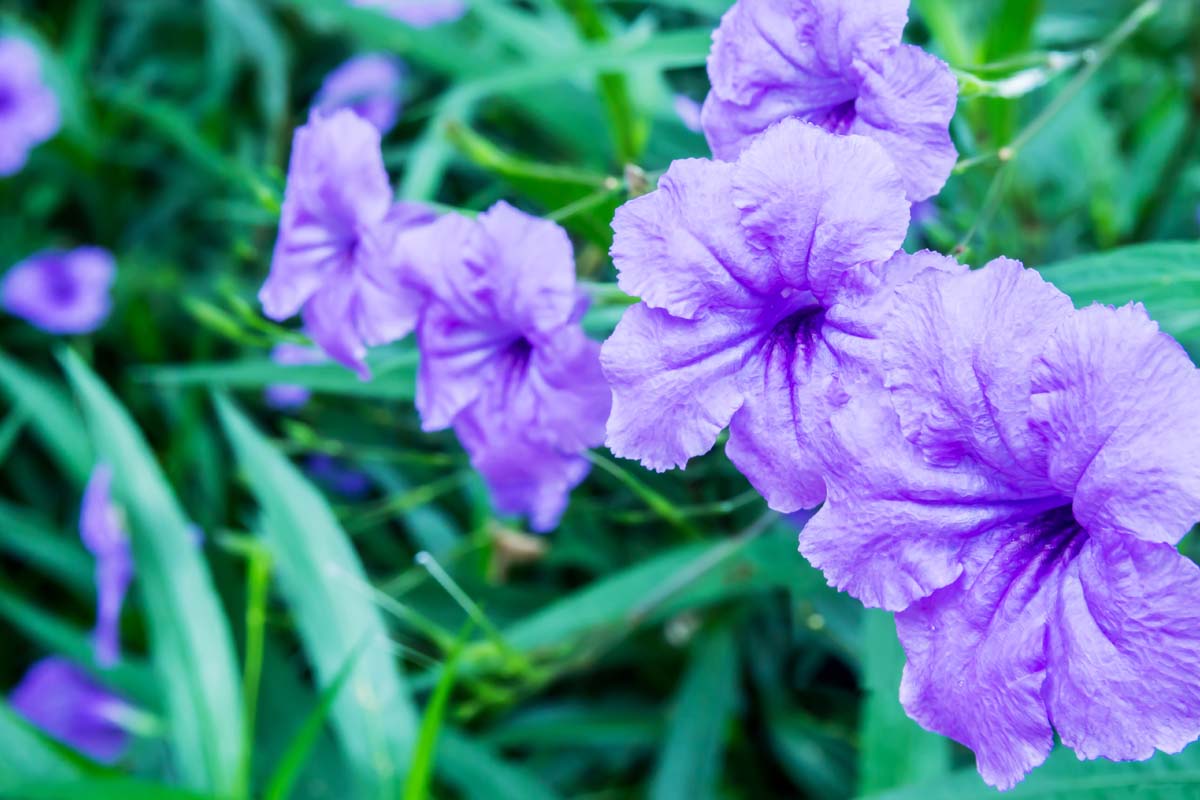} &
        \includegraphics[width=48px, height=48px]{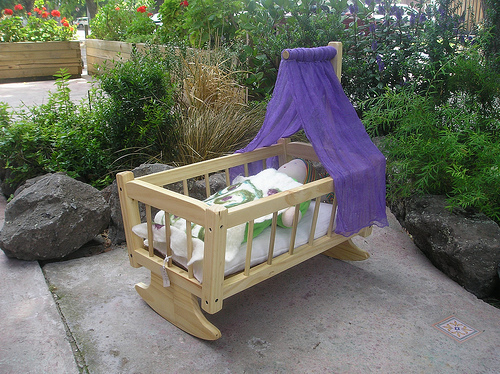} & \includegraphics[width=48px, height=48px]{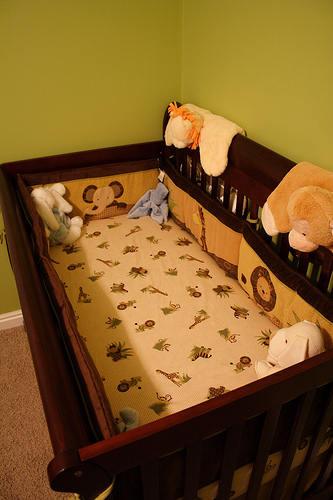}
        \\
        \footnotesize{\textcolor{darkgreen}{Blanket Flower}} & \footnotesize{\textcolor{darkgreen}{Mexican Petunia}} & \footnotesize{\textcolor{darkgreen}{Infant Bed}} & \footnotesize{\textcolor{darkgreen}{Infant Bed}} \\
        
        \footnotesize{\textcolor{red}{Gaillardia}} & \footnotesize{\textcolor{red}{Ruellia}} & \footnotesize{\textcolor{red}{Cradle}} & \footnotesize{\textcolor{red}{Crib}} \\
        \bottomrule
    \end{tabular}
    \caption{The tables shows the images with ground truth labels in green and predicted labels in red. Even though the predicted labels are sementically similar, SLAC model do not able to match them. In these situations TLAC model is effective.}
    \label{tab:images-2step}
\end{table}

Consider the images in Table \ref{tab:images-2step}. Dataset labels are shown in green, and our model's predicted labels are in red. For the first two flower images, our SLAC model predicted their scientific names, ``Gaillardia'' and ``Ruellia,'' instead of the common names ``Blanket Flower'' and ``Mexican Petunia.'' Similarly, for the third and fourth images, the ground truth label was ``Infant Bed'', but our model predicted ``Cradle'' and ``Crib''. While these predictions are technically correct, our SLAC model struggled to identify the semantic equivalence between these terms. However, our TLAC model successfully corrected these mismatches, demonstrating the advantage of iterative refinement in certain scenarios.

%%%%%%%%%%%%%%%%% Domain Generalization Table %%%%%%%%%%%%%%
\begin{table}[]
    % \centering
    \resizebox{\linewidth}{!}{
    \begin{tabular}{c c c c c}
    \toprule
         Methods & ImageNetV2 & ImageNet-S  & ImageNet-R \\
         \midrule
         CLIP  & 60.83 & 46.15 & 73.96 \\
         CoOp  & 64.20 & 47.99 & 75.21 \\
         Co-CoOp & 64.07 & 48.75 & 76.18 \\
         Maple  & 64.07 & 49.15 & 76.98 \\
         MMA & 64.33 & 49.13 & 77.32 \\
         \bottomrule
         \textbf{Ours (SLAC)} & \textcolor{blue}{67.93} & \textcolor{blue}{66.11} & \textcolor{blue}{89.91}  \\

         \textbf{Ours (TLAC)} & \textbf{69.21} & \textbf{68.20} & \textbf{90.82}  \\

         \bottomrule
    \end{tabular}
    }
    \caption{The table presents the results of our method compared to state-of-the-art few-shot methods. Higher scores better. The best result is displayed in bold, while the second-highest in blue.}
    \label{tab:domain-gen}
\end{table}

\subsection*{Domain Generalization}

Table \ref{tab:domain-gen} presents the performance of our approach compared to previous methods \cite{clip, coop, maple} on domain generalization datasets. While previous methods were trained on ImageNet, our approach was training-free. Our model achieved a 3.6\% improvement over the state-of-the-art MMA model on ImageNetV2. Similarly, on ImageNet-S,  and ImageNet-R, our approach yielded significant gains of 16.96\%, and 12.59\%, respectively.

\section{Ablation Study}
\subsection{Effect of Prompt}

Prior work \cite{maple, he2024does} has demonstrated the influence of prompt design on LLM/LMM accuracy, a phenomenon we also observed in our model. This effect is particularly pronounced for domain-specific datasets. The reason is LLMs/LMMs are highly sensitive to prompt wording as they rely on pattern recognition and context. A well-crafted prompt serves as a guide, offering the necessary context and examples to effectively ``program'' the model. This leads to more accurate outputs, especially when dealing with complex, domain-specific datasets that require nuanced understanding. To investigate this, we evaluated our model on a general-purpose dataset (Caltech101) and a domain-specific dataset (StanfordCars). Table \ref{tab:ablation-caltech} shows the impact of varying prompts on Caltech101.  Asking the LMM ``which object is present in the image'' yielded lower accuracy compared to the more specific prompt ``which specific object is present in the image.''  Further improvement was observed when prompting for a description: ``what is the specific type of object present in the image,'' achieving an accuracy of 95.63\%, compared to 94.43\% and 94.98\% with the previous prompts. While the accuracy differences are not significant, optimized prompting consistently improves performance.

\begin{table}[]
    % \centering
    % \resizebox{\linewidth}{!}{
    \begin{tabular}{|p{6.6cm}|c|}
        \hline
        \textbf{Prompt} & \textbf{Acc.} \\ \hline
         which object is present in the image. & 94.43 \\ \hline
         which specific object is present in the image. & 94.98 \\ \hline
         what is the specific type of object present in the image. & 95.63 \\ \hline
    \end{tabular}
    % }
    \caption{Table showing the effect of different prompts on image classification accuracy of Caltech101 dataset. Acc. means Accuracy. Higher accuracy is better.}
    \label{tab:ablation-caltech}
\end{table}

On the domain-specific StanfordCars dataset, prompt engineering significantly impacts accuracy. As shown in Table \ref{tab:ablation-stanfordcars}, the generic prompt ``which object is present in the image'' yielded an accuracy of 84.55\%, while the domain-specific prompt ``which car is present in the image'' improved accuracy to 88.83\%.  Further specifying the prompt to ``which specific car is present in the image'' resulted in an even higher accuracy of 90.20\%. This demonstrates the importance of domain-specific prompts for domain-specific datasets.

\begin{table}[]
    % \centering
    % \resizebox{\linewidth}{!}{
    \begin{tabular}{|p{6.6cm}|c|}
        \hline
        \textbf{Prompt} & \textbf{Acc.} \\ \hline
         which object is present in the image. & 84.55 \\ \hline
         which car is present in the image. & 88.83 \\ \hline
         which specific car is present in the image. & 90.20 \\ \hline
    \end{tabular}
    % }
    \caption{Table showing the effect of different prompts on image classification accuracy of StanfordCars dataset. Acc. means Accuracy. Higher accuracy is better.}
    \label{tab:ablation-stanfordcars}
\end{table}

%%%%%%%%%%%%%%%%%%%%%%%%%%%%%%%%%

\subsection{Difference between Gemini Pro and Flash}
\begin{figure}[]
\begin{center}
   \includegraphics[width=1.0\linewidth]{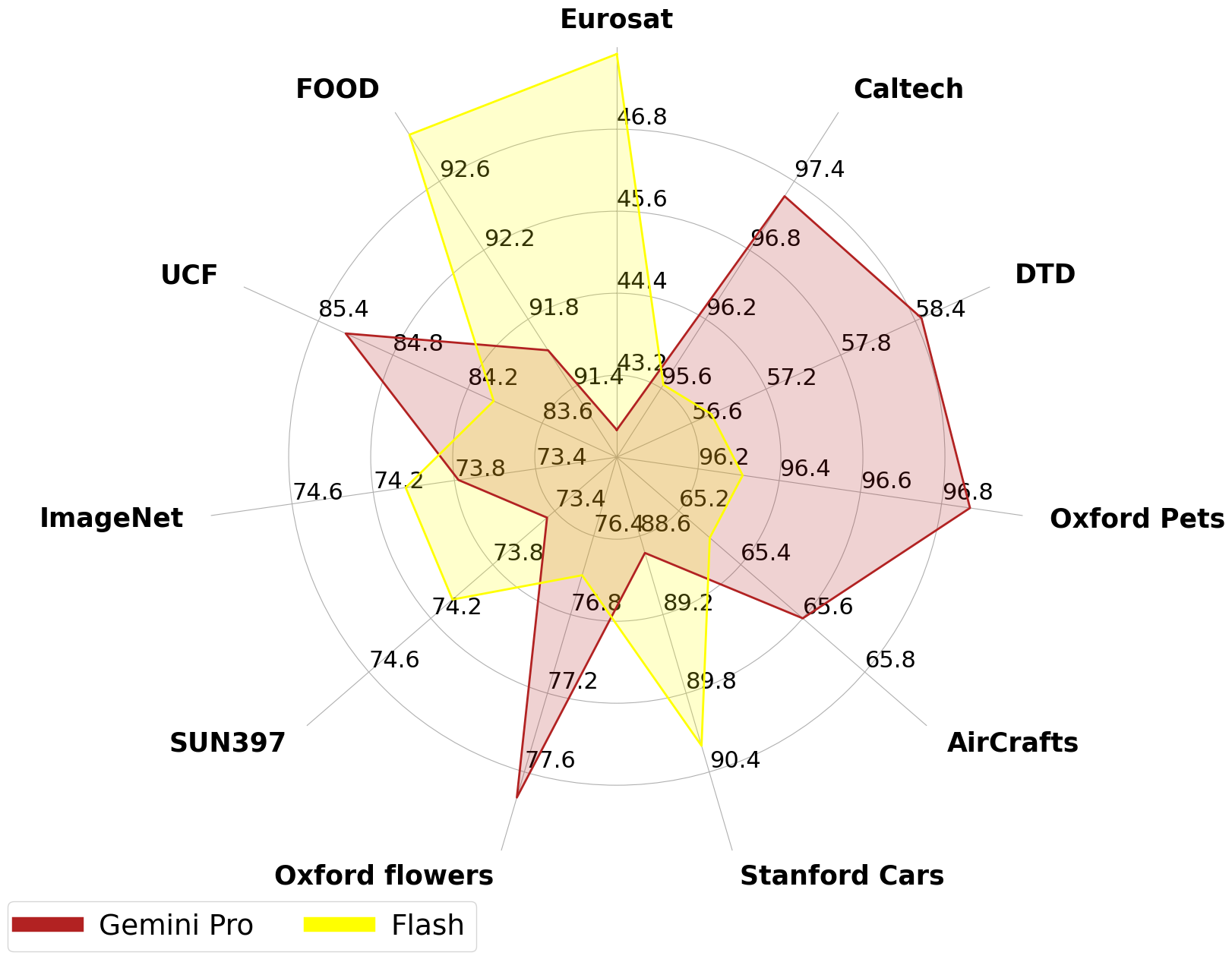}
\end{center}
   \caption{The figure illustrates the difference of accuracy between Gemini Pro and Gemini Flash LMMs in our model.}
\label{fig:pro-vs-flash}
\end{figure}

Figure \ref{fig:pro-vs-flash} compares the performance of our model using Gemini Pro and Gemini Flash LMM backbones in our SLAC model.  While some datasets show slight performance variations between the two, the overall difference is not significant, and both achieve comparable results. Given the cost-effectiveness and speed advantages of Gemini Flash, it offers a compelling alternative to the more expensive and relatively slower Gemini Pro.

\subsection{Effect of VLM}
\begin{figure}[]
\begin{center}
   \includegraphics[width=1.0\linewidth]{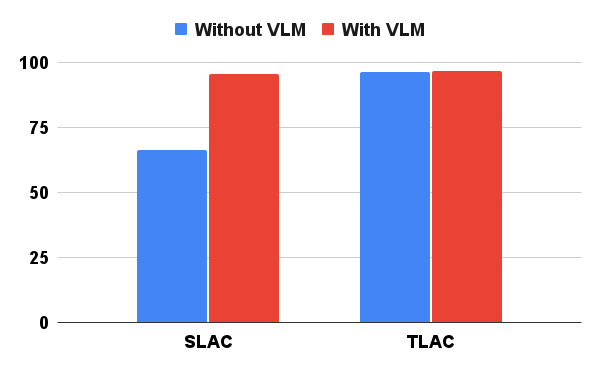}
\end{center}
   \caption{This figure demonstrates the impact of integrating a VLM, CLIP, on the accuracy of our proposed architecture.}
\label{fig:with-and-without-vlm}
\end{figure}

To evaluate the impact of the CLIP on our model architecture, we conducted experiments using the Caltech101 dataset and the Gemini Flash configuration. As illustrated in Figure \ref{fig:with-and-without-vlm}, we observed a significant accuracy difference in the SLAC model. Without CLIP, utilizing only the LMM, Gemini Flash, resulted in an accuracy of 66.27\%. Conversely, incorporating CLIP yielded a substantial improvement, reaching 96.62\%. This disparity highlights the critical role of CLIP in the performance of SLAC model. In contrast, the TLAC model exhibited a minimal accuracy variation, achieving 96.18\% without CLIP and 96.51\% with CLIP, suggesting CLIP's reduced significance in this particular architecture.

\subsection{Adding classes in prompt}
While class names were not included in our standard Gemini prompts, our experiment demonstrates a performance boost when they are provided. For instance, accuracy of our SLAC model on the OxfordPets dataset improved from 96.48\% to 98.27\% with the inclusion of class names. To maintain a realistic evaluation, reflecting scenarios where class names are often unavailable, we did not incorporate them in our experiments.

\section{Discussion and Future Direction}
This work demonstrates the advantages of using pre-trained Large Multimodal Models (LMMs) for image classification. A key contribution is our approach that requires no training or fine-tuning on new datasets and thus offering strong generalization capabilities. This generalization stems from the web-scale data used to train LMMs. While prior work has explored using CLIP, trained on 400 million image-text pairs \cite{clip}, which in its original form often struggles to achieve high accuracy on various datasets, even with few-shot fine-tuning. In contrast, LMMs leverage significantly larger datasets. Consequently, our method achieves higher average accuracy on 13 datasets as compared to previous state-of-the-art training-free approaches. Similarly, our method also achieved better accuracy than previous few-shot methods on 9 of 11 standard image classification datasets and on all four domain generalization datasets, without any training or fine-tuning.

This work explores image classification with LMMs, achieving promising results. However, the potential for data contamination presents a critical challenge. Given the scale and diversity of LMM training data, it is possile that these models have encountered images similar to, if not portions of, commonly used benchmarks. To address this, future research should prioritize developing robust LMM evaluation methodologies, including the creation of large-scale, diverse, held-out datasets distinct from common training data.  Such datasets are crucial for assessing true generalization ability of LMMs.

\section{Limitations}
While the aforementioned advantages are notable, our method also presents certain limitations. A primary concern pertains to the LMM model, specifically whether the observed accuracy is attributable to our unique model configuration or the inherent power of the Gemini architecture. Furthermore, the potential for data contamination exists, raising concerns that the LMM model may have been exposed to the datasets employed for evaluation. To mitigate the first limitation, future studies could investigate the performance of different LMM models to isolate the contribution of our specific configuration from that of the underlying backbone. Regarding the second limitation, the development of a novel benchmark dataset, previously unseen by the LMM, would be a valuable contribution. We recommend that future research endeavors focus on these areas.

Another limitation is that LMMs tend to show lower accuracy on highly specific datasets. For example, our model underperforms fine-tuned CLIP on OxfordPets because fine-tuning allows models to learn dataset-specific correlations, a significant advantage on specialized datasets like OxfordPets. Similarly, we observe this trend with EuroSAT, a low-resolution (64x64 pixel) aerial image dataset with 10 scene classes (e.g., River). Consequently, datasets with high specialization (e.g., OxfordPets) or low image resolution (e.g., EuroSAT) currently favor fine-tuned CLIP models over LMMs.

%For instance, it underperforms compared to fine-tuned CLIP models on the OxfordPets dataset. This performance gap can be attributed to the differing training paradigms. LMMs, trained on extensive and diverse datasets, develop more generalized object representations. Conversely, fine-tuned models learn dataset-specific correlations, providing an advantage on specialized datasets like OxfordPets. Consequently, highly specific datasets tend to yield lower accuracy with LMMs. A similar trend is observed with the EuroSAT dataset, which comprises low-resolution (64x64 pixel) aerial images categorized into 10 scene classes, such as River. The limited image resolution poses a challenge for LMMs in accurately identifying or describing the scene. Fine-tuned models, however, demonstrate superior accuracy on this dataset due to their training on class-specific features. Therefore, datasets characterized by either high specialization (e.g., OxfordPets) or low image resolution (e.g., EuroSAT) currently favor fine-tuned CLIP models over LMMs.

\section{Literature Review}
\subsection{Vision Language Models}
Recently, vision language models (VLMs), which are trained on both vision and text simultaneously, have gained significant popularity due to their impressive zero-shot and few-shot capabilities. Single modality models, such as vision or language models, are trained in isolation, which results in a modality gap between them. Recent VLMs include CLIP \cite{clip}, ALIGN \cite{ALIGN}, FILIP \cite{yao2021filip}, and Florence \cite{yuan2021florence}. These VLMs have been trained on large amounts of data in a self-supervised manner. Despite being trained on large datasets, these VLMs still face challenges in achieving state-of-the-art results on downstream tasks.

\subsection{Efficient Transfer Learning for VLMs}
Efficient transfer learning methods, such as prompt learning and adapter-based approaches, have been introduced to adapt Vision-Language Models (VLMs) to downstream tasks. These methods add a small number of parameters to the pre-trained VLM, updating only these newly introduced parameters during fine-tuning while keeping the original VLM parameters frozen. Text prompts, consisting of sentence-based instructions can be either handcrafted or learned during training (a process known as ``Prompt Learning''). Originally developed in Natural Language Processing (NLP) \cite{lester2021power, li2021prefix, liu2021p}, prompt learning has been successfully adopted in VLMs \cite{cocoop, coop, zhu2023prompt}. Extending beyond text prompts, some methods \cite{rao2022denseclip, tsimpoukelli2021multimodal} have explored prompts within the VLM's visual encoder. Maple \cite{maple} introduced multimodal prompt learning, training prompts in both the language and visual branches. CoOp \cite{coop} refines continuous prompt vectors within the language module to improve CLIP's few-shot transfer learning performance. Addressing CoOp's limitations in generalizing to unseen categories, Co-CoOp \cite{cocoop} conditions prompts on input images. Further enhancing prompt learning, \cite{proDA} proposes optimizing multiple prompt sets by learning their underlying distribution.

In addition to prompt learning, an alternative approach involves integrating lightweight modules, termed adapters, into VLMs for adaptation to downstream tasks \cite{clip-adapter, zhang2022tip-adapter}. CLIP-Adapter \cite{clip-adapter} fine-tunes vision-language models by incorporating feature adapters on either the visual or language branch. AdaptFormer \cite{chen2022adaptformer}, an approach for Vision Transformers (ViTs), introduces lightweight modules to enhance transferability without modifying pre-trained weights. Multi-Modal Adapter \cite{yang2024mma} aligns the representations learned by adapters in the vision and text branches by adding multimodal adapters.

\subsection{Training-Free Methods}

Tip-Adapter \cite{zhang2022tip-adapter} is a training-free method that adapts CLIP for few-shot learning by adding a non-parametric adapter. SuS-X \cite{udandarao2023sus} adapts CLIP for training-free image classification by dynamically creating support sets from category names and using them within a zero-shot framework to guide predictions. These methods employ dataset class features for training-free inference.  Alternative approaches utilize class descriptions. For examle, CuPL \cite{CuPL} combines a Large Language Model (LLM) with CLIP to generate customized prompts for image classification. By using an LLM, CuPL creates numerous category-specific prompts with detailed visual descriptions, enabling better discrimination between similar classes in a zero-shot manner. Meta-Prompting for Visual Recognition (MPVR) \cite{MPVR} uses a system prompt, task description, and a fixed in-context example to guide an LLM in generating visual style-infused query templates.  %These templates are then populated with class labels to create category-specific VLM prompts, which are used to build an ensemble of zero-shot classifiers.

\subsection{Large MultiModal Models (LMMs)}
Following the rise of LLMs, Large Multimodal Models (LMMs) have become a prominent research area. Large multimodal models are developed in two stages: pre-training, where the model is trained on a massive dataset using self-supervised tasks like next-word prediction; and post-training, where it is fine-tuned to follow instructions. GPT-4 \cite{achiam2023gpt4}, developed by OpenAI, is a large multimodal model that can process both image and text inputs to generate text outputs. Claude 3 \footnote{https://claude.ai/} is a multimodal model that accepts multiple input modalities, including text, tables, graphs, and photos. Developed by Google AI, Gemini \footnote{https://gemini.google.com/app} \cite{gemini1.0, gemini1.5} is a cutting-edge multimodal model capable of understanding and generating content across different modalities, including text, code, images, and other data types. Gemini comes in two versions: Flash and Pro. Flash is designed for speed and cost efficiency while Pro, is engineered for superior accuracy.

The large multimodal models mentioned above are closed-source, meaning their precise architecture and training data are not publicly disclosed. In contrast, open-source models make their architecture and trained weights publicly available. Pixtral-12B \cite{agrawal2024pixtral} is a 12-billion-parameter multimodal language model. Qwen2-VL \cite{wang2024qwen2} introduces a dynamic resolution mechanism for processing images. LLaVA (Large Language and Vision Assistant) \cite{llava} is an end-to-end trained LMM that combines CLIP with the Vicuna \cite{chiang2023vicuna}. MiniGPT-4 \cite{zhu2023minigpt} aligns a frozen visual encoder with a frozen large language model (Vicuna) to replicate GPT-4's advanced multimodal abilities. %, demonstrating that proper alignment of visual features with a powerful LLM can achieve capabilities like detailed image description and website generation from sketches. 
Llama \cite{llama3.1} is a new family of language models. While Llama 3 \cite{llama3} and 3.1 \cite{llama3.1} were language models, Llama 3.2 introduced multimodal capabilities. %These models achieve comparable quality to leading models like GPT-4 across various tasks. 

This paper proposes using a Large Multimodal Model (LMM) in a training-free fashion, eliminating the need for LLM fine-tuning due to its pre-training on extensive volume of data. Specifically, we utilize Gemini for inference due to its readily accessible API through the Google Cloud Platform (GCP). The initial \$300 GCP credit provided at signup enabled us to use Gemini across all our datasets at virtually no cost. In contrast, open-source models like Llama require access to significant GPU resources for inference. Unfortunately we lack the compute resources to evaluate our scheme using Llama. Our work presents results using both Gemini Pro and Gemini Flash.

\section{Conclusion}
Standard pre-trained Vision-Language Models (VLMs) like CLIP often require fine-tuning to achieve high accuracy on downstream tasks such as image classification. Full model fine-tuning is computationally expensive, leading to the exploration of efficient alternatives like prompt learning (which introduces a small number of learnable parameters) and adapter layers (small multilayer perceptrons). However, both methods require fine-tuning for each new dataset, demanding significant computational resources. In contrast, we propose leveraging Large Multimodal Models (LMMs) such as Gemini for image classification. Trained on massive datasets, LMMs have demonstrated impressive language and vision understanding capabilities. Their extensive pre-training eliminates the need for further fine-tuning, allowing direct application to new datasets without additional training. Our results demonstrate that LMMs achieve significant improvements on large general datasets like ImageNet, as well as domain-specific datasets such as Flowers102, FGVCAircraft, and StanfordCars. Future works can explore other LMMs, both open-source and closed-source, and compare their performance with that of Gemini.

%ucf-train-2
%dtd-train-2
%imagenet-train-2
%imagenet-v2-2

{
    \small
    \bibliographystyle{ieeenat_fullname}
    \bibliography{main}
}


\end{document}